\documentclass{ecai2010}
\usepackage{times}
\usepackage{graphicx}
\usepackage{latexsym}
\usepackage{amsthm}

\usepackage{xspace}

\newcommand{\myOmit}[1]{}

\begin{document}
\newcommand{\set}{\mathcal}
\newcommand{\myset}[1]{\ensuremath{\mathcal #1}}

\renewcommand{\theenumii}{\alph{enumii}}
\renewcommand{\theenumiii}{\roman{enumiii}}
\newcommand{\figref}[1]{Figure \ref{#1}}
\newcommand{\tref}[1]{Table \ref{#1}}
\newcommand{\myldots}{\ldots}

\newtheorem{mydefinition}{Definition}
\newtheorem{mytheorem}{Proposition}
\newtheorem*{myexample}{Running Example}{\bf}{\it}
\newtheorem*{myexampletwo}{Example}{\bf}{\it}
\newtheorem{mytheorem1}{Theorem}
\newcommand{\myproof}{\noindent {\bf Proof:\ \ }}
\newcommand{\myqed}{\mbox{$\Box$}}
\newcommand{\myend}{\mbox{$\clubsuit$}}

\newcommand{\mymod}{\mbox{\rm mod}}
\newcommand{\mymin}{\mbox{\rm min}}
\newcommand{\mymax}{\mbox{\rm max}}
\newcommand{\range}{\mbox{\sc Range}}
\newcommand{\roots}{\mbox{\sc Roots}}
\newcommand{\myiff}{\mbox{\rm iff}}
\newcommand{\alldifferent}{\mbox{\sc AllDifferent}}
\newcommand{\permutation}{\mbox{\sc Permutation}}
\newcommand{\disjoint}{\mbox{\sc Disjoint}}
\newcommand{\cardpath}{\mbox{\sc CardPath}}
\newcommand{\CARDPATH}{\mbox{\sc CardPath}}
\newcommand{\common}{\mbox{\sc Common}}
\newcommand{\uses}{\mbox{\sc Uses}}
\newcommand{\lex}{\mbox{\sc Lex}}
\newcommand{\usedby}{\mbox{\sc UsedBy}}
\newcommand{\nvalue}{\mbox{\sc NValue}}
\newcommand{\slide}{\mbox{\sc CardPath}}
\newcommand{\sliden}{\mbox{\sc AllPath}}
\newcommand{\SLIDE}{\mbox{\sc CardPath}}
\newcommand{\circularslide}{\mbox{\sc CardPath}_{\rm O}}
\newcommand{\among}{\mbox{\sc Among}}
\newcommand{\mysum}{\mbox{\sc MySum}}
\newcommand{\amongseq}{\mbox{\sc AmongSeq}}
\newcommand{\atmost}{\mbox{\sc AtMost}}
\newcommand{\atleast}{\mbox{\sc AtLeast}}
\newcommand{\element}{\mbox{\sc Element}}
\newcommand{\gcc}{\mbox{\sc Gcc}}
\newcommand{\gsc}{\mbox{\sc Gsc}}
\newcommand{\contiguity}{\mbox{\sc Contiguity}}
\newcommand{\PRECEDENCE}{\mbox{\sc Precedence}}
\newcommand{\assignnvalues}{\mbox{\sc Assign\&NValues}}
\newcommand{\linksettobooleans}{\mbox{\sc LinkSet2Booleans}}
\newcommand{\domain}{\mbox{\sc Domain}}
\newcommand{\symalldiff}{\mbox{\sc SymAllDiff}}
\newcommand{\alldiff}{\mbox{\sc AllDiff}}
\newcommand{\doublelex}{\mbox{\sc DoubleLex}}

\newcommand{\slidingsum}{\mbox{\sc SlidingSum}}
\newcommand{\MaxIndex}{\mbox{\sc MaxIndex}}
\newcommand{\REGULAR}{\mbox{\sc Regular}}
\newcommand{\regular}{\mbox{\sc Regular}}
\newcommand{\precedence}{\mbox{\sc Precedence}}
\newcommand{\STRETCH}{\mbox{\sc Stretch}}
\newcommand{\SLIDEOR}{\mbox{\sc SlideOr}}
\newcommand{\NAE}{\mbox{\sc NotAllEqual}}
\newcommand{\mytheta}{\mbox{$\theta_1$}}
\newcommand{\mysigma}{\mbox{$\sigma_2$}}
\newcommand{\mysigmatwo}{\mbox{$\sigma_1$}}

\newcommand{\todo}[1]{{\tt (... #1 ...)}}

\newcommand{\dpsb}{DPSB}

\title{Symmetries of Symmetry Breaking Constraints}

\author{George Katsirelos\institute{CRIL-CNRS, Lens, France, 
email: gkatsi@gmail.com}
\and 
Toby Walsh\institute{NICTA and UNSW, Sydney, Australia,
email: toby.walsh@nicta.com.au}}


\maketitle
\bibliographystyle{ecai2010}

\begin{abstract}
Symmetry is an important feature of many
constraint programs. We show that \emph{any} problem symmetry
acting on a set of symmetry breaking constraints 
can be used to break symmetry.
Different symmetries pick out different solutions in each symmetry class. 
This simple but powerful idea can be used in a number
of different ways. We describe one application within
model restarts, a search technique designed to reduce
the conflict between symmetry breaking and the branching
heuristic. In model restarts, we restart search periodically with a 
random symmetry of the symmetry breaking constraints. 
Experimental results show that this symmetry breaking technique is
effective in practice on some standard benchmark problems. 
\end{abstract}

\section{INTRODUCTION}

Symmetry occurs in many real world problems. 
For instance, certain machines in a scheduling problem
might be identical. If we have a valid schedule, 
we can permute these machines and still have a valid schedule. 
We typically need to factor such symmetry 
out of the search space to be able to find solutions efficiently. 
One popular way to deal with symmetry
is to add constraints which
eliminate symmetric
solutions (see, for instance, 
\cite{puget:Sym,ssat2001,ffhkmpwcp2002,llconstraints06,wecai2006,waaai2008}). 
Such symmetry breaking is usually simple to implement 
\cite{fhkmwcp2002,fhkmwaij06} 
and is often highly efficient and effective in practice.
Even for problems with many symmetries, 
a small number of symmetry breaking 
constraints can often eliminate
much or all of the symmetry. 
But where do such symmetry breaking constraints 
come from? We show here that we can apply
symmetry to symmetry breaking constraints
themselves to generate potentially new symmetry
breaking constraints. 

There are a number of applications of this
simple but powerful idea. We 
give one application in the area of
restart based methods. Restarting has 
proven a powerful technique to deal
with branching mistakes in backtracking
search \cite{gskaaai98}. 
One problem with posting symmetry breaking constraints
is that they pick out particular solutions in each
symmetry class, and branching heuristics may conflict
with this choice.
{\em Model restarts} is a technique to deal with this 
conflict \cite{hpsycp08}. 
We periodically restart search with 
a new model containing
different symmetry breaking constraints. 
Our idea of applying symmetry to symmetry breaking
constraints provides a systematic method to generate
different symmetry breaking constraints to be
used within model restarts. Different symmetries
pick out different solutions in each symmetry
class. Restarting search with a different
symmetry of the symmetry breaking constraints may therefore permit
symmetry breaking constraints to be posted
that do not conflict with the branching heuristic. 
Our experimental results show that model
restarts is indeed effective at reducing the conflict
between branching heuristics and symmetry breaking.

\section{SYMMETRY}

We consider two common types of symmetry
(see \cite{cjjpsconstraints06} for more discussion). 
A \emph{variable symmetry}
is a permutation of the variables
that preserves solutions. 
Formally,
a variable symmetry 
is a bijection $\sigma$ on the
indices of variables such that if $X_1=d_1, \ldots, X_n=d_n$ is a solution
then $X_{\sigma(1)}=d_1, \ldots, X_{\sigma(n)}=d_n$ is also. 
A \emph{value symmetry} 
is a permutation of the values
that preserves solutions. Formally,
a value symmetry 
is a bijection $\theta$ on the
values such that if $X_1=d_1, \ldots, X_n=d_n$ is a solution
then $X_1=\theta(d_1), \ldots, X_n=\theta(d_n)$ is also. 
In \cite{getree}, these are called \emph{global value symmetries}
as their action on values is the same for all variables. 
Symmetries can more generally act 
on both variables and values. 
Our results apply also to such symmetries. 
As the inverse of a symmetry and the identity
are symmetries, the set of symmetries 
forms a group under composition. 

We will use a simple running example
which has a small number of 
symmetries. 
This example will demonstrate that we can use
symmetry itself to pick out different solutions
in each symmetry class. 
%
\begin{myexample}
A magic square is a labelling of a $n$ by $n$
square with the numbers 
1 to $n^2$ so that the sums of each row, column
and diagonal are equal
(prob019 in CSPLib \cite{csplib}). 
The most-perfect magic square problem
is to find a magic square 
in which every 2 by 2 square has the same
sum, and in which all pairs of integers $n/2$ apart on
either diagonal have the same sum. 
We model this as a CSP 
with
$X_{i,j}=k$ iff the square $(i,j)$ 
contains $k$. 
One solution for $n=4$ is:
\begin{eqnarray}\label{ms1}
&
\begin{tabular}{|c|c|c|c|} \hline
14 & 11 & 5 & 4 \\ \hline
1 & 8 & 10 & 15 \\ \hline
12 & 13 & 3 & 6 \\ \hline
7 & 2 & 16 & 9 \\ \hline
\end{tabular}
&
\end{eqnarray}
This is one of the oldest
known most-perfect magic squares,
dating from a 10th century 
temple engraving 
in Khajuraho, India. 

This 
problem has several 
symmetries.
First, there are the 8 
symmetries of the square:
the identity mapping,
the rotations 90$^\circ$ clockwise,
180$^\circ$ and 270$^\circ$,
and the reflections in the 
vertical, horizontal and diagonal
axes.
For example, 
applying the symmetry $\sigma_v$ that reflects 
the square in its vertical axis 
to (\ref{ms1}) 
gives a symmetric solution: 
\begin{eqnarray} \label{ms2}
&
\begin{tabular}{|c|c|c|c|} \hline
4 & 5 & 11 & 14 \\ \hline
15 & 10 & 8 & 1 \\ \hline
6 & 3 & 13 & 12 \\ \hline
9 & 16 & 2 & 7 \\ \hline
\end{tabular}
&
\end{eqnarray}

The problem also has a value symmetry $\theta_{inv}$
that inverts values, mapping $i$ onto $n^2+1-i$. 
For instance, applying $\theta_{inv}$ to (\ref{ms1}), 
generates another symmetric solution:
\begin{eqnarray} \label{ms3}
&
\begin{tabular}{|c|c|c|c|} \hline
3 & 6 & 12 & 13 \\ \hline
16 & 9 & 7 & 2 \\ \hline
5 & 4 & 14 & 11 \\ \hline
10 & 15 & 1 & 8 \\ \hline
\end{tabular}
&
\end{eqnarray}

We can also combine the value and variable
symmetries. 
For example, if we 
apply the composition of the last two
symmetries (that is, $\theta_{inv} \circ \sigma_v$) to 
(\ref{ms1}), we reflect the solution in the vertical axis
and invert all values giving another symmetric solution:
\begin{eqnarray} \label{ms4}
&
\begin{tabular}{|c|c|c|c|} \hline
13 & 12 & 6 & 3 \\ \hline
2 & 7 & 9 & 16 \\ \hline
11 & 14 & 4 & 5 \\ \hline
8 & 1 & 15 & 10 \\ \hline
\end{tabular}
&
\end{eqnarray}
Note that (\ref{ms4}) is itself the reflection of
(\ref{ms3}) in the vertical axis. 
The problem thus has 
16 symmetries in total. 
\myend 
\end{myexample}

\section{SYMMETRY BREAKING}

One common way to deal with symmetry 
is to add constraints to eliminate 
symmetric solutions \cite{puget:Sym}. 
We shall show that we can in fact break symmetry by posting
a \emph{symmetry} of any such symmetry breaking constraints.
%
%
\begin{myexample}
Consider again the most-perfect magic square problem. 
To eliminate the symmetries of the square,
we can post the constraints:
\begin{eqnarray} \label{symbreak1}
X_{1,1} < \mymin(X_{1,n},X_{n,1},X_{n,n}),
& \ \ &
X_{1,n} < X_{n,1}
\end{eqnarray}
These ensure that the smallest
corner is top left, and
the bottom left corner is smaller
than the top right. This eliminates
all degrees of freedom to rotate
and reflect the square.
Note that (\ref{symbreak1})
eliminates solutions (\ref{ms1})
and (\ref{ms4}) but leaves 
their reflections (\ref{ms2})
and (\ref{ms3}). 

To eliminate the value symmetry $\theta_{inv}$
and any rotations or reflections of it, 
we can post:
\begin{eqnarray} \label{symbreak2}
& X_{1,1} < n^2+1 - \mymax(X_{1,1},X_{1,n},X_{n,1},X_{n,n}) &
\end{eqnarray}
This ensures that the smallest
corner in the magic square is smaller than 
the smallest corner in any rotation
or reflection of the inversion of the solution. 
Note that the smallest corner cannot 
\emph{equal} the inversion of any corner since
the problem constraints of a most-perfect
magic square ensure that the inversion
of any corner lies inside the square (in fact,
on the diagonal). This symmetry breaking
constraint eliminates
(\ref{ms2}) but leaves (\ref{ms3}). 
Thus, of the four symmetric solutions given
earlier, only (\ref{ms3})
satisfies (\ref{symbreak1}) and (\ref{symbreak2}). 
\myend \end{myexample}

One of our main observations is that {\em any symmetry} 
acting on a set of symmetry breaking constraints
will itself break the symmetry in a problem. 
Different symmetries pick out different
solutions in each symmetry class. 
To show this, we need to consider
the action of a symmetry on a set of symmetry 
breaking constraints. 
Symmetry is often defined as acting on assignments, 
mapping solutions to solutions. 
We need to lift this definition to constraints. 
The action of a variable symmetry on
a constraint changes the variables on which the constraint
acts. More precisely, a variable symmetry 
$\sigma$
applied to the constraint 
$C(X_j,\ldots,X_k)$
gives $C(X_{\sigma(j)},\ldots,X_{\sigma(k)})$.
The action of a value symmetry is also easy to
compute. A value symmetry 
$\theta$ applied to the constraint 
$C(X_j,\ldots,X_k)$
gives $C(\theta(X_j),\ldots,\theta(X_k))$. 

\begin{myexample}
To illustrate how we can break symmetry
with the symmetry of a set of symmetry breaking
constraints, we shall construct symmetries of
(\ref{symbreak1}) and (\ref{symbreak2}). 

Consider the symmetry 
$\sigma_{v}$ that
reflects the square in the vertical axis
mapping $X_{1,1}$ 
onto $X_{n,1}$ (and vice versa),
and $X_{1,n}$ onto $X_{n,n}$ (and vice versa). 
If we apply $\sigma_{v}$ 
to (\ref{symbreak1}) we get:
\begin{eqnarray} \label{symbreak3}
X_{n,1} < \mymin(X_{n,n},X_{1,1},X_{1,n}), & \ \ & 
X_{n,n} < X_{1,1} 
\end{eqnarray}
These new symmetry breaking
constraints ensure that the smallest
corner is now top right, and
the bottom right corner is smaller
than the top left. This again eliminates
all degrees of freedom to rotate
and reflect the square.
Note that (\ref{symbreak3})
eliminates solutions (\ref{ms2})
and (\ref{ms3}) but leaves 
(\ref{ms1}) and
(\ref{ms4}). This is the opposite
of posting (\ref{symbreak1}) which
would leave (\ref{ms2}) and (\ref{ms3}),
but eliminate
(\ref{ms1}) and
(\ref{ms4}).

If we apply $\sigma_{v}$ to (\ref{symbreak2}), 
we again get a constraint that breaks
the value symmetry $\theta_{inv}$
and any rotations or reflections of it:
\begin{eqnarray} \label{symbreak4}
& X_{n,1} < n^2+1 - \mymax(X_{n,1},X_{n,n},X_{1,1},X_{1,n}) &
\end{eqnarray}
This again ensures that the smallest
corner in the magic square is smaller than 
the smallest corner in any rotation
or reflection of the inversion of the solution. 
This eliminates
(\ref{ms1}) but leaves (\ref{ms4}). 
Thus, of the four symmetric solutions given
earlier, only (\ref{ms4})
satisfies the symmetry $\sigma_v$ 
of (\ref{symbreak1}) and (\ref{symbreak2}). 

We can also break symmetry with any other symmetry of 
the symmetry breaking constraints. 
For instance, 
if we apply $\theta_{inv} \circ \sigma_{v}$ to 
(\ref{symbreak1}) 
we get the constraints:
\begin{eqnarray*} 
& n^2+1-X_{n,1} < \mymin(n^2+1-X_{n,n},n^2+1-X_{1,1},n^2+1-X_{1,n}) & \\
& n^2+1-X_{n,n} < n^2+1-X_{1,1}  & 
\end{eqnarray*}
These simplify to: 
\begin{eqnarray*}
& X_{n,1} > \mymax(X_{n,n},X_{1,1},X_{1,n}) & \\
& X_{n,n} > X_{1,1} & 
\end{eqnarray*}
Similarly, if we apply $\theta_{inv} \circ \sigma_v$ to 
(\ref{symbreak2}) 
we get the constraint:
\myOmit{
\begin{eqnarray*} 
& n^2+1-X_{n,1} < n^2+1 - \mymax(n^2+1-X_{n,1},n^2+1-X_{n,n},n^2+1-X_{1,1},n^2+1-X_{1,n}) & 
\end{eqnarray*}}
\begin{eqnarray*}
& X_{n,1} > n^2+1 - \mymin(X_{n,1},X_{n,n},X_{1,1},X_{1,n})  & 
\end{eqnarray*}
The three constraints ensure that the largest corner
is top right, the bottom right is larger
than the top left, and the 
largest corner is larger than the
largest corner in any rotation or reflection 
of the inversion of the solution.
This again prevents
us from rotating, reflecting or
inverting any solution.
Of the four symmetric solutions given
earlier, only (\ref{ms2}) satisfies
$\theta_{inv} \circ \sigma_v$ of (\ref{symbreak1}) and (\ref{symbreak2}). 
We see therefore that different symmetries of the symmetry
breaking constraints pick out
different solutions in each symmetry class. 
\myend \end{myexample}


\section{THEORETICAL RESULTS}

The running example illustrates 
that we can break symmetry with a symmetry of 
a set of symmetry breaking constraints. We will
prove that this holds in general. Our aim is
to show: 
\begin{quote}
{\em Any symmetry acting on a set of symmetry
breaking constraints itself breaks
symmetry. Different symmetries pick
out different solutions in each symmetry class.
}
\end{quote}
We will consider the action of a symmetry on
the solutions and symmetries of a set of 
symmetry breaking constraints. We will also study the action
of a symmetry on the soundness and completeness of a set of
symmetry breaking constraints, 
the representative solutions picked out
by the symmetry breaking constraints and the symmetries
that are eliminated. Finally we will consider what
symmetries can be found within a set of
symmetry breaking constraints. 

\subsection{Symmetry and satisfiability}

We start with the action of
a symmetry on the satisfiability of
a set of constraints. 
This simple
result is used in some of the later
proofs. 
We write $\sigma(C)$ for
the result of applying the symmetry $\sigma$ to 
the set of constraints $C$. 

\begin{mytheorem}
\label{thm0} 
For any symmetry $\sigma$, 
a set of constraints $C$ is satisfiable
iff $\sigma(C)$ is satisfiable. 
\end{mytheorem}
\myproof
Suppose $C$ is satisfiable. 
Then there exists a satisfying assignment $A$ of $C$.
By considering the action of a 
symmetry on a set of constraints,
we can see that $\sigma(A)$ satisfies $\sigma(C)$. Thus
$\sigma(C)$ is satisfiable. The proof reverses
easily. 
\myqed

\subsection{Symmetry and solutions}

We next consider the action of a symmetry
on the solutions of a set of (possibly symmetry breaking)
constraints. We write $sol(C)$ for the set of solutions to 
the set of constraints $C$. 

\begin{mytheorem}
For any symmetry $\sigma$ and set of constraints $C$:
\label{thm1} 
$$sol(\sigma(C))=\sigma(sol(C))$$
\end{mytheorem}
\myproof
Consider any solution $A \in sol(\sigma(C))$. 
We view a solution as a set
of assignments. Then $\sigma(C) \cup A$ is satisfiable. 
As $\sigma$ is a bijection, there exists
a unique $B$ such that $A=\sigma(B)$. 
Thus $\sigma(C) \cup \sigma(B)$ is satisfiable. 
Hence $\sigma(C \cup B)$ is satisfiable. 
By Proposition \ref{thm0}, 
$C \cup B$ is satisfiable. That is, $B \in sol(C)$. 
Thus $\sigma(B) \in \sigma(sol(C))$. 
Hence $A \in \sigma(sol(C))$. The proof reverses
directly. 
\myqed

On the other hand, if we apply a symmetry
to a set of constraints with that same symmetry,
we do not change the set of solutions.

\begin{mytheorem}
\label{thm2} 
If $\sigma$ is a symmetry of a set of constraints $C$ 
then $$sol(C)=\sigma(sol(C))$$ 
\end{mytheorem}
\myproof
Consider any $A \in \sigma(sol(C))$. Then
there exists $B \in sol(C)$
such that $A=\sigma(B)$. Since $\sigma$ is
a symmetry of $C$, $\sigma(B) \in sol(C)$. 
That is $A \in sol(C)$. The proof reverses
directly. 
\myqed

\subsection{Symmetries under symmetry}

The action of a symmetry
on a set of constraints also
does not change the symmetries of
those constraints. 

\begin{mytheorem}
If $\Sigma$ is a symmetry group of a set of constraints $C$ 
then $\Sigma$ is also a symmetry group of $\sigma(C)$
for any $\sigma \in \Sigma$. 
\end{mytheorem}
\myproof
Consider any solution $A$ of $\sigma(C)$ and any $\tau \in \Sigma$.
Since $\sigma \in \Sigma$, by Proposition \ref{thm2},
$sol(\sigma(C)) = sol(C)$. 
Thus $A \in sol(C)$. 
As $\tau \in \Sigma$, $\tau(A) \in sol(C)$. 
Hence $\tau(A) \in sol(\sigma(C))$. 
It follows that $\tau$ is a symmetry
of $\sigma(C)$. 
\myqed

\subsection{Symmetry and soundness}

An important property of a set of symmetry
breaking constraints is its soundness. 
For a problem with symmetries $\Sigma$,
a set of symmetry breaking constraints is
{\em sound} iff it leaves at least one solution 
in each symmetry class. 
All the symmetry breaking
constraints used in our running example
are sound. 
The action of a symmetry on a set of
symmetry breaking constraints leaves
their soundness unchanged. 

\begin{mytheorem}[Soundness]
Given a set of symmetries $\Sigma$ of $C$, 
if $S$ is a sound set of symmetry breaking
constraints for $\Sigma$
then $\sigma(S)$ for any $\sigma \in \Sigma$
is also a sound set of symmetry breaking constraints
for $\Sigma$. 
\end{mytheorem}
\myproof
Consider any
$A \in sol(C \cup S)$ and
any $\sigma \in \Sigma$. 
Now $A \in sol(C)$
and $A \in sol(S)$. But as $\sigma$ is 
a symmetry of $C$, $\sigma(A) \in sol(C)$. 
Since $A \in sol(S)$, it follows from Proposition \ref{thm1}
that $\sigma(A) \in sol(\sigma(S))$.
Thus, $\sigma(A) \in sol(C \cup \sigma(S))$. 
Hence, there is at least one solution 
left by $\Sigma(S)$ in every symmetry class of $C$.
That is, $\sigma(S)$ is a sound set of
symmetry breaking constraints for $\Sigma$. 
\myqed

\subsection{Symmetry and completeness}

A set of symmetry breaking constraints may also be complete. 
For a problem with symmetries $\Sigma$,
a set of symmetry breaking constraints is 
{\em complete} iff it leaves at most one
solution in each symmetry class.
The action of a symmetry on a set of
symmetry breaking constraints leaves
their completeness unchanged. 

\begin{mytheorem}[Completeness]
Given a set of symmetries $\Sigma$ of $C$, 
if $S$ is a complete set of symmetry breaking
constraints for $\Sigma$
then $\sigma(S)$ for any $\sigma \in \Sigma$
is also a complete set of symmetry breaking constraints
for $\Sigma$. 
\end{mytheorem}
\myproof
Consider any $\sigma \in \Sigma$ and
$A \in sol(C \cup \sigma(S))$. 
Now $A \in sol(C)$
and $A \in sol(\sigma(S))$. 
But as $\sigma$ is 
a symmetry of $C$, so is 
$\sigma^{-1}$. Hence
$\sigma^{-1}(A) \in sol(C)$. 
Since $A \in sol(\sigma(S))$, it follows from Proposition \ref{thm1}
that $\sigma^{-1}(A) \in sol(S)$.
Thus
$\sigma^{-1}(A) \in sol(C \cup S)$. 
Hence, there is at most one solution 
left by $\sigma(S)$ in every symmetry class of $C)$.
That is, $\sigma(S)$ is a complete set of
symmetry breaking constraints for $\Sigma$. 
\myqed

\subsection{Representative solutions}

Different symmetries of the symmetry breaking constraints 
pick out different solutions in each symmetry class.
In fact, we can pick out any solution we like
by choosing the appropriate symmetry of a set of
symmetry breaking constraints. 

\begin{mytheorem}
Given a symmetry group $\Sigma$ of a set of 
constraints $C$, a sound set $S$ of symmetry breaking
constraints, 
and any solution $A$ of $C$, 
then there is a symmetry $\sigma \in \Sigma$ 
such that $A \in sol(C \cup \sigma(S))$. 
\end{mytheorem}
\myproof
Since the set of symmetry breaking constraints
$S$ is sound, it leaves at least one
solution (call it $B$) in the same symmetry class as $A$. That is,
$B \in sol(C \cup S)$. Hence
$B \in sol(C)$ and $B \in sol(S)$. 
As $A$ and $B$ are in the same symmetry class, there exists
a symmetry $\sigma$ in $\Sigma$ with $A=\sigma(B)$.
Since $B \in sol(S)$, it follows from Proposition \ref{thm1}
that $\sigma(B) \in sol(\sigma(S))$.
That is, $A \in sol(\sigma(S))$. 
As $B \in sol(C)$ and $\sigma \in \Sigma$, 
it follows that $\sigma(B) \in sol(C)$. 
That is, $A \in sol(C)$. 
Hence $A \in sol(C \cup \sigma(S))$.
\myqed

We will use this result in the second half
of the paper where we consider
the conflict between symmetry breaking
constraints and branching heuristics. 
We will exploit the fact that 
whatever solution the branching heuristic is going towards, 
there exists a symmetry of the symmetry breaking
constraints which does not conflict with this.

\subsection{Symmetries eliminated}

In certain cases, a set of symmetry breaking constraints
completely eliminates a symmetry. 
%
We say that a set of symmetry breaking constraints $S$ 
{\em breaks} a symmetry 
$\sigma$ of a problem $C$ iff there exists a solution $A$ of $C \cup S$ 
such that $\sigma(A)$ is not a solution of $C \cup S$,
and {\em eliminates} a symmetry 
$\sigma$ iff for each solution $A$ of $C \cup S$,
$\sigma(A)$ is not a solution of $C \cup S$. 
Similarly, $S$ {breaks (eliminates)} a set of symmetries
$\Sigma$ iff $S$ breaks (eliminates) each $\sigma \in
\Sigma$. 

It is not hard to
see that a sound and complete set of symmetry breaking constraints
eliminates every non-identity symmetry. 
\myOmit{
\begin{mytheorem}
If $S$ is a sound and complete set of symmetry breaking
constraints for the symmetries $\Sigma$ of $C$, 
then $S$ 
eliminates every non-identity symmetry in $\Sigma$. 
\end{mytheorem}
\myproof
Consider any non-identity symmetry $\sigma$ in $\Sigma$. 
Since $S$ is sound and complete, it leaves a single
solution in each symmetry class. Consider any
such solution $A$. Then $\sigma(A)$ is eliminated
by $S$. Hence, $S$ eliminates the symmetry $\sigma$. 
\myqed}
However, there are 
symmetry breaking constraints 
which break a particular
symmetry but do not \emph{eliminate} it. 

\myOmit{
\begin{myexampletwo}
Consider finding 3 by 3 0/1 matrices
which contain 4 non-zero entries. 
This problem has symmetry of interchangeable rows and columns.
We can break this with a symmetry breaking
constraint that lexicographically 
orders the rows and columns \cite{ffhkmpwcp2002}. 
Take the symmetry that
maps row 1 onto row 2, 
row 2 onto row 3, 
row 3 onto row 1,
column 1 onto column 3,
column 2 onto column 1,
and column 3 onto column 2.
Symmetry breaking eliminates this symmetry
as it permits the following solution:
$$
\begin{array}{ccc}
0 & 0 & 1 \\
0 & 1 & 0 \\
1 & 1 & 0
\end{array}
$$
But the symmetry of this solution is ruled
out as the row and columns are
not lexicographically ordered:
$$
\begin{array}{ccc}
1 & 0 & 1 \\
0 & 1 & 0 \\
1 & 0 & 0
\end{array}
$$
On the other hand, symmetry breaking does not
{completely} eliminate this symmetry 
as the following solution demonstrates:
$$
\begin{array}{ccc}
0 & 0 & 1 \\
0 & 1 & 1 \\
1 & 0 & 0
\end{array}
$$
This has rows and columns that are
lexicographically ordered, but
the symmetry of this solution also has 
rows and columns that are
lexicographically ordered:
$$
\begin{array}{ccc}
0 & 0 & 1 \\
0 & 1 & 0 \\
1 & 1 & 0
\end{array}
$$
\end{myexampletwo}
}
\begin{myexample}
Consider again one of the symmetry breaking 
constraints in (\ref{symbreak3}):
\begin{eqnarray} \label{symbreak5}
& X_{1,n} < X_{n,1} & 
\end{eqnarray}
This eliminates the 
symmetry that reflects 
the magic square in its trailing
diagonal. If we take
any solution which satisfies
(\ref{symbreak5}), then
any reflection of this solution in the
trailing diagonal is removed by (\ref{symbreak5}).
Note that (\ref{symbreak5}) also breaks
the symmetry $\sigma_{90}$ (90$^\circ$ clockwise rotation) 
since (\ref{ms2}) satisfies
(\ref{symbreak5}) but $\sigma_{90}$ of 
(\ref{ms2}) does not. However, 
(\ref{symbreak5}) does not eliminate
$\sigma_{90}$. For example, both $\sigma_{270}$ of 
(\ref{ms2}) and $\sigma_{90}$ of this solution
satisfy (\ref{symbreak5}). 
\myend 
\end{myexample}

Applying a symmetry
to a set of symmetry breaking constraints 
changes the solutions in each symmetry
class accepted by the symmetry breaking
constraints. However,
it does not change the symmetries broken or eliminated
by the symmetry breaking constraints. 

\begin{mytheorem}
Given a problem $C$ with a symmetry group $\Sigma$, 
if $S$ breaks (eliminates) $\Sigma$
then $\sigma(S)$ breaks (eliminates) $\Sigma$
for any $\sigma \in \Sigma$.
\end{mytheorem}
\myproof
Suppose $S$ breaks $\Sigma$.
Consider any symmetry $\tau \in \Sigma$.
Then there exists a solution $A$ of $C \cup S$
such that $\tau(A)$ is not a solution of $C \cup S$.
As $A$ is a solution of $C$ and $\tau \in \Sigma$, $\tau(A)$ is
a solution of $C$. Hence
$\tau(A)$ is not a solution of $S$.
By Proposition \ref{thm1}, 
$\sigma(\tau(A))$ is not a solution
of $\sigma(S)$.
Since $\Sigma$ is a group, it is closed
under composition. Thus $\tau \circ \sigma \circ \tau^{-1} \in \Sigma$.
Hence, as $A \in sol(C)$, $\tau^{-1}(\sigma(\tau(A))) \in sol(C)$. 
Thus there is a solution of $C$, namely 
$\tau^{-1}(\sigma(\tau(A))) \in sol(C)$
such that the symmetry $\tau$ of this 
(which equals $\sigma(\tau(A))$) is not a solution
of $\sigma(S)$. Hence, $\sigma(S)$ breaks
$\tau$. The proof for when $S$ eliminates
$\Sigma$ follows similar lines. 
\myqed

\subsection{Symmetries of symmetry breaking constraints}

We have discussed the action of a symmetry on
a set of symmetry breaking constraints. But what
can we say about the symmetries of a set of symmetry
breaking constraints? 

\begin{mytheorem}
If the symmetry breaking constraints 
$S$ break the symmetries $\Sigma$ in the
set of constraints $C$ 
then $S$ does not have any symmetry in $\Sigma$. 
\end{mytheorem}
\myproof
Consider any symmetry $\sigma \in \Sigma$. 
Suppose $S$ has this symmetry. 
Since $S$ breaks $\sigma$
there exists a solution 
$A$ of $C \cup S$ such that 
$\sigma(A)$ is not a solution of $S$. 
Hence, $A$ is a solution of $S$ but
$\sigma(A)$ is not. Thus, $S$ does not have any symmetry in $\Sigma$. 
\myqed

The reverse does not necessarily hold. There exist
constraints which lack a symmetry which it
is sound to post but which do not break
that symmetry. 

\begin{myexample}
Consider again the most-perfect magic squares problem. 
Consider the following constraint: 
$$X_{1,1}+X_{n,n}= n^2+1 \rightarrow X_{1,1}<X_{n,n}$$
This does not have any variable symmetry since
we cannot interchange the two variables. 
However, this constraint does not break
any variable symmetry since $X_{1,1}+X_{n,n} \neq n^2+1$
in every most-perfect magic square. 
\myend 
\end{myexample}

\section{MODEL RESTARTS}

The idea that the symmetries of symmetry breaking
constraints can themselves be used to break symmetry
can be used in several different ways. 
We consider here an application of this idea 
for tackling the conflict between branching
heuristics and symmetry breaking constraints. 
Symmetry breaking picks out particular solutions
in each symmetry class and these may not be the
same solutions towards which branching heuristics 
are directing search. 
Heller {\it et al.} propose using {\it model restarts}
\cite{hpsycp08} to tackle this conflict. Backtracking search is
restarted periodically, using a new model 
which contains different symmetry breaking constraints.
By posting different symmetry breaking 
constraints, we hope at some point
for the branching heuristic and symmetry breaking
not to conflict. 
Heller {\it et al.} do not, however, provide
a general method to generate different symmetry 
breaking constraints after each restart. 

Our observations that any symmetry
acting on a set of symmetry breaking constraints
can be used to break symmetry, and that different
symmetries pick out different solutions, 
provide us with \emph{precisely} the tool we need
to perform model restarts to \emph{any} domain (and not just
to the domain of interchangeable variables and values studied
in \cite{hpsycp08}). 
When we restart
search, we simply post a different symmetry of 
the symmetry breaking constraints. 
We experimented with several possibilities.
The simplest was to choose a symmetry
at random from the symmetry group. 
We also tried various heuristics
like using the symmetry most consistent or
most inconsistent with 
previous choices of the branching heuristic.
However, we observed the best performance of
model restarts with a random choice of symmetry so we 
only report results here with such a choice.
This is also algorithmically simple since computer
algebra packages like GAP provide efficient algorithms 
for computing a random element of a group given a set of
generators for the group. 

\begin{myexample}
We consider the simple problem
of finding a magic square of order 5.
The following table gives the amount of
search needed to find such a magic
square when posting one of the rotational 
symmetries of the symmetry breaking
constraints and the default branching
heuristic that labels variables in a fixed
order. We encoded the problem in BProlog
running on a Pentium 4 3.2 GHz
processor with 3GB of memory. 
With magic squares in general, we cannot guarantee
that the inverse of the smallest corner
is not itself a corner value. We therefore
relax the strict inequality in (\ref{symbreak2}),
replacing it by a non-strict inequality. 

{\rm
\begin{center}
\begin{tabular}[h]{|c||r|r|} \hline
Symmetry posted 
& Backtracks & Time to solve/s \\ 
of  (\ref{symbreak1}) and (\ref{symbreak2})  & & \\ \hline
$\sigma_{id}$ & 658 & 0.02 \\
$\sigma_{90}$ & 17,143 & 0.36 \\
$\sigma_{180}$ & 315,267 & 5.60 \\
$\sigma_{270}$ & 18,808,974 & 408.85 \\ \hline
\end{tabular}
\end{center}
}

We see that
the different symmetries of the symmetry breaking
constraints interact differently with the branching
heuristic. 
Model restarts will help overcome this conflict. 
Suppose we restart search every 1,000 backtracks
and choose to post
at random one of these symmetries
of (\ref{symbreak1}) and (\ref{symbreak2}). 
Let $t$ be the average number of branches
to find a solution. There is $\frac{1}{4}$ chance
that the first restart will post $\sigma_{id}$ of
(\ref{symbreak1}) and (\ref{symbreak2}).
In this situation,
we find a solution after 658 backtracks.
Otherwise we post
one of the other symmetries of 
(\ref{symbreak1}) and (\ref{symbreak2}). 
We then
explore 1,000 backtracks, reach the cutoff and fail to 
find a solution. As each restart is independent,
we restart and explore on average another $t$ more branches. 
Hence:
\begin{eqnarray*}
t & = & \frac{1}{4} 658 + \frac{3}{4}(1000+t)
\end{eqnarray*}
Solving for $t$ gives $t=3,658$. Thus, using model
restarts, 
we take
just 3,658 backtracks on average to solve the problem.
\myend 
\end{myexample}

Note that posting random symmetries of the symmetry
breaking constraints is not 
equivalent to fixing the symmetry breaking and
randomly branching. Different symmetries of the 
symmetry breaking constraints interact in different ways 
with the problem constraints. Although the problem
constraints are themselves initially symmetrical,
branching decisions quickly break the symmetries
that they have. 

\section{EXPERIMENTAL RESULTS}

\begin{table*}[htb]
  \scriptsize
  \centering
  \begin{tabular}[ht]{|l||rr|rr||rr|rr||rr|rr|}
    \hline
    & 
    \multicolumn{4}{|c||}{Static posting} &
    \multicolumn{4}{|c||}{Model restarts} &
    \multicolumn{4}{|c|}{SBDS}\\
    &
    \multicolumn{2}{|c|}{Lex} &
    \multicolumn{2}{|c||}{Random} &
    \multicolumn{2}{|c|}{Lex} &
    \multicolumn{2}{|c||}{Random} &
    \multicolumn{2}{|c|}{Lex} &
    \multicolumn{2}{|c|}{Random} 
    \\
    & 
    opt &
    t/b &
    opt &
    t/b &
    opt &
    t/b &
    opt &
    t/b &
    opt &
    t/b &
    opt &
    t/b
    \\
    \hline \hline
1& \textbf{13} & \textbf{0.11} & \textbf{13} &114.99 & \textbf{13} &1.13 & \textbf{13} &4.99 &13 * &0&14 * &270.42 \\
&  & \textbf{387} &  &424 K &  &2087&  &13 K &  &3639 K &  &4309 K \\
2& \textbf{10} & \textbf{0.03} & \textbf{10} &354.72 & \textbf{10} &2.67 &- &- & \textbf{10} &0.04 & \textbf{10} &0.77 \\
&  & \textbf{31} &  &1364 K &  &6503&  &- &  &229&  &4539\\
3&12 * &0.02 &- &- & \textbf{12} & \textbf{3.44} & \textbf{12} &137.55 &12 * &0&12 * &0.02 \\
&  &2052 K &  &- &  & \textbf{7958} &  &386 K &  &3178 K &  &2744 K \\
4& \textbf{14} &6.12 & \textbf{14} &15.91 & \textbf{14} & \textbf{5.14} & \textbf{14} &22.7 &14 * &0&20 * &0.01 \\
&  &25 K &  &76 K &  & \textbf{13 K} &  &54 K &  &3040 K &  &3188 K \\
5& \textbf{16} & \textbf{0.3} &- &- & \textbf{16} &26.16 & \textbf{16} &6.06 &16 * &0&16 * &0.01 \\
&  & \textbf{730} &  &- &  &38 K &  &12 K &  &3193 K &  &3083 K \\
6& \textbf{13} &247.85 &- &- & \textbf{13} & \textbf{115.41} & \textbf{13} &203.25 &14 * &0&15 * &90.27 \\
&  &1001 K &  &- &  & \textbf{410 K} &  &693 K &  &3313 K &  &3676 K \\
7& \textbf{8} &0.03 &- &- & \textbf{8} &0.72 & \textbf{8} &0.75 & \textbf{8} & \textbf{0.02} & \textbf{8} &0.12 \\
&  & \textbf{60} &  &- &  &1980&  &1794&  &119&  &684\\
8& \textbf{17} & \textbf{0.1} &- &- & \textbf{17} &20.35 & \textbf{17} &97.9 &17 * &0&17 * &0.01 \\
&  & \textbf{170} &  &- &  &59 K &  &284 K &  &2820 K &  &2634 K \\
9& \textbf{20} & \textbf{0.22} & \textbf{20} &0.48 & \textbf{20} &111.06 & \textbf{20} &10.56 &20 * &0.01 &20 * &0.01 \\
&  & \textbf{387} &  &1103&  &172 K &  &15 K &  &3490 K &  &3442 K \\
10& \textbf{8 *} &0.01 & \textbf{8 *} &25.29 & \textbf{8 *} &1.78 & \textbf{8 *} &5.46 & \textbf{8 *} & \textbf{0} & \textbf{8 *} &0.02 \\
&  &4003 K &  &4612 K &  &3460 K &  &3510 K &  &3135 K &  &4309 K \\
    \hline \hline
  \end{tabular}
  \caption{Static symmetry breaking constraints vs Model restarts and
    SBDS on Graph Coloring problems. 
``opt'' is the quality of the solution
found (* indicates optimality was not proved), ``t'' is the runtime in 
seconds, ``b'' is the number of backtracks. The best method for a problem
instance is in {\bf bold font}.}
  \label{tab:results}
\end{table*}

\begin{table*}[htb]
  \scriptsize
  \centering
  \begin{tabular}{|c||ll|ll||ll|ll||ll|ll|}
    \hline
    Instance & 
    \multicolumn{4}{|c||}{Static constraints} &
    \multicolumn{4}{|c||}{Model Restarts} &
    \multicolumn{4}{|c|}{SBDS} \\
    &
    \multicolumn{2}{|c|}{Lex} &
    \multicolumn{2}{|c||}{Random} &
    \multicolumn{2}{|c|}{Lex} &
    \multicolumn{2}{|c||}{Random} &
    \multicolumn{2}{|c|}{Lex} &
    \multicolumn{2}{|c|}{Random}
    \\
    & 
    t &
    b &
    t &
    b &
    t &
    b &
    t &
    b &
    t &
    b &
    t &
    b 
    \\
    \hline\hline
3-3-4-5& \textbf{0.01} & \textbf{176} & \textbf{0.01} &246&0.02 &271&0.2 &3504&0.06 &2060& \textbf{0.01} &318\\
3-4-6-5&0.02 &599&0.04 &726&0.04&703&0.59 &7340&0.03 &855& \textbf{0.01} & \textbf{465} \\
4-3-3-3& \textbf{0} & \textbf{34} &0.01 &141&0.12 &1570&0.08 &1013&0.01 &51& \textbf{0} &57\\
4-3-4-5&0.02 &326&0.14 &2392&0.28 &3821&0.14 &1629&0.16 &4177& \textbf{0.01} & \textbf{42} \\
4-3-5-4& \textbf{0} &91&0.06 &1204&0.12 &1497&0.23&4055& \textbf{0} &52& \textbf{0} & \textbf{51} \\
4-4-4-5&0.07 &1350&0.03 &314&1.77 &17 K &6.88 &98 K & \textbf{0} & \textbf{41} &0.04 &930\\
4-4-5-4&0.02 &504&0.27 &3956&1.01 &8941&1.03 &10 K & \textbf{0.01} & \textbf{208} &0.04 &790\\
4-6-4-5&45.43 &763 K &-&-&0.27 & 2612 &0.67 &5104&6.59 &111 K & \textbf{0.04} & \textbf{629} \\
5-3-3-4& \textbf{0.01} & \textbf{190} &0.06 &937&6.64 &75 K &3.6 &63 K &0.14 &3113&0.02 &382\\
5-3-4-5& \textbf{0.03} & \textbf{527} &1.51 &21 K &4.34 &45 K &1.32 &13 K &1.16 &26 K &0.12 &2109\\
6-3-4-5&0.25 &4538&2.84 &40 K &3.9 &44 K &9.06 &106 K &5.92 &114 K & \textbf{0.06} & \textbf{1380} \\
    \hline
  \end{tabular}
  \caption{Static symmetry breaking constraints vs Model restarts and SBDS on EFPA problems.
``t'' is the runtime in 
seconds, ``b'' is the number of backtracks. The best method for a problem
instance is in {\bf bold font}.}
  \label{tab:r2}
\end{table*}

Our experiments are designed to test two hypotheses. 
The first hypothesis is that model restarts is less
sensitive to branching heuristics than
posting static symmetry breaking constraints. 
%
We test this hypothesis by 
using either a lexicographic value ordering
(which does not conflict with the symmetry 
breaking constraints) or random value ordering,
while using min-domain for variable ordering.
We expect that model restarts will show
smaller variation between the two value
ordering heuristics.
The second hypothesis is that model restarts
will often explore a smaller search tree than dynamic 
methods like SBDS due to propagation
of the symmetry breaking constraints.
We tested two domains
that exhibit two different kinds of symmetry: partial
variable and value interchangability, and
row and column symmetry.

We limit our comparison of dynamic methods
to SBDS. Whilst there is a specialized
dynamic symmetry breaking
method for interchangeable variables and
values, experiments in \cite{hpsycp08}
show that this is several orders of magnitude
slower than static methods. 
We also do not compare to methods
such as GE-trees \cite{getree}, as this method is
limited to value symmetry and does not deal with
the variable/value symmetries in our domains. 
Finally, we used SBDS to break
just generators of the symmetry group
as breaking the full symmetry group 
quickly exhausted available memory. 
%
%
%
We implemented model restarts and SBDS
as well as all the static symmetry
breaking constraints
in Gecode 2.2.0, and ran
all experiments on a 4-core Intel Xeon 5130 
with 4MB of L2 cache running at 2GHz.

The first set of experiments uses random graph coloring
problems generated in the same way as the previous 
experimental study in \cite{llwycp07}. 
All values in this
model are interchangeable. 
In addition, we introduce variable symmetry 
by partitioning variables into 
interchangeable sets of size at most 8. We
randomly connect the vertices within each partition with either a
complete graph or an empty graph, and choose each option with equal
probability. Similarly, between any two partitions there is equal
probability that the partitions are completely connected or
independent. Results for graphs with 40
vertices are shown in Table 1. 


The second set of experiments uses
Equidistant Frequency Permutation Array (EFPA)
problem \cite{hmmncp09}.
Given the parameters $v, q, \lambda, d$, 
the objective is to find $v$ codewords of length
$q\lambda$, such that each word contains
exactly $\lambda$ occurences of each
of $q$ symbols and each pair of words
have Hamming distance $d$. 
%
Our model has
both row and column symmetry. We implement
model restarts by randomly choosing a permutation
of the rows and columns and posting 
lexicographic ordering constraints on the rows and
columns of the resulting matrix.

The 
results
support both our hypotheses. 
Although the model restarts method is 
not necessarily the best method for any given
instance, its performance is most robust. 
The variability of the runtimes between the 
lex value ordering and random value ordering
is much smaller for model restarts, as well as for
SBDS. This suggests that in domains where
the branching heuristic interacts more strongly
with the problem constraints, model restarts
will be more robust than static symmetry breaking constraints.
Our second hypothesis, that model restarting tends to explore a
smaller search tree than SBDS is also supported by
the graph coloring results. SBDS was unable to
prove optimality in all but one instance in graph
coloring. 
These results confirm the
findings of \cite{hpsycp08}.
In EFPA on the other hand, 
there are enough solutions
in these instances that applying 
little symmetry breaking 
with a random value ordering 
seems to be the best strategy.
However, on the harder instances such as 4-6-4-5,
using model restarts
resolves the conflict between
the branching heuristic and the
symmetry breaking constraints
and achieves parity with SBDS.

\section{OTHER RELATED WORK}

Crawford {\it et al.} proposed a general method
to break symmetry statically using lex-leader 
constraints \cite{clgrkr96}. 
Like other static
methods, the posted 
constraints
pick out in advance a particular solution in each
symmetry class. Unfortunately, this may conflict with the 
solution sought by the branching heuristic. 
There are a number of symmetry breaking methods
proposed to deal with this conflict. 
For example, dynamic symmetry breaking methods like
SBDS posts symmetry breaking constraints
dynamically during search \cite{sbds}. 
Another dynamic method for breaking symmetry is SBDD 
\cite{fahle1}. This checks if a node of the
search tree is symmetric to some previously explored 
node. A weakness of such dynamic methods is that
we get little or no propagation
on the symmetry breaking constraints.
It has been shown that propagation between the problem constraints
and the static symmetry breaking constraints can reduce
search exponentially \cite{wcp07}. 

Jefferson {\it et al,} have
proposed GAPLex, a hybrid method
that combines together static and dynamic symmetry
breaking \cite{gaplex}. 
However, GAPLex is limited to dynamically posting 
lexicographical ordering constraints, and to 
searching with a fixed variable ordering. 
%
Puget has proposed ``Dynamic Lex'', a hybrid method that
dynamically posts static symmetry breaking
constraints during search which works with dynamic
variable ordering heuristics \cite{puget2003}.
This method adds 
symmetry breaking constraints dynamically during search that
are compatible with the current partial assignment. 
In this way, the first solution found during tree search
is not removed by symmetry breaking. 
Dynamic Lex needs to compute the stabilizers of the current partial
assignment. This requires a 
graph isomorphism problem to be solved at each node of the search tree. 
Whilst Dynamic Lex works with dynamic variable
ordering heuristics, it assumes that
values are tried in order. 
Finally Dynamic Lex is limited to posting
lexicographical ordering constraints. 
A comparison 
with Dynamic Lex is interesting but
challenging. 
For instance, Heller {\it et al.}  \cite{hpsycp08} did not compare
model restarts with Dynamic Lex, arguing:
\begin{quote}
{\em ``It is not clear how this method [Dynamic Lex] can be generalized, 
though, and for the case
of piecewise variable and value symmetry, no method with similar 
properties is known yet.''}
\end{quote}

\section{CONCLUSIONS}

We have considered the action of symmetry
on symmetry breaking constraints. We proved that {any symmetry} 
applied to a set of symmetry breaking constraints
gives a (possibly new) set of symmetry breaking
constraints that break the same symmetries. 
In addition, we proved that different symmetries of the
set of symmetry breaking constraints will
pick out different solutions in each symmetry class. 
We used these observations to help tackle
the conflict between symmetry breaking 
and branching heuristics.
In particular, we applied these ideas to 
{\em model restarts} 
\cite{hpsycp08}. In this search technique, 
we periodically restart search with a new model
which contains a random
symmetry of the symmetry breaking constraints. 
Experimental results show that this helps 
keep many of the benefits of posting static
symmetry breaking constraints 
whilst reducing the conflict between symmetry 
breaking and the branching heuristic. 
There are other potential applications of these
ideas. For example, we are currently developing 
methods for dynamically posting a symmetry of the symmetry
breaking constraints which does not conflict
with the branching heuristic. In the longer
term, we would like to 
exploit symmetries of the nogoods
learnt during search. Nogoods
are themselves just constraints.
We can therefore consider symmetries
acting on them. 

\bibliography{/Users/twalsh/Documents/biblio/a-z,/Users/twalsh/Documents/biblio/a-z2,/Users/twalsh/Documents/biblio/pub,/Users/twalsh/Documents/biblio/pub2}

\begin{thebibliography}{10}

\bibitem{cjjpsconstraints06}
D.~Cohen, P.~Jeavons, C.~Jefferson, K.E. Petrie, and B.M. Smith, `Symmetry
  definitions for constraint satisfaction problems', {\em Constraints}, {\bf
  11}(2--3),  115--137, (2006).

\bibitem{clgrkr96}
J.~Crawford, M.~Ginsberg, G.~Luks, and A.~Roy, `Symmetry breaking predicates
  for search problems', in {\em 5th Int. Conf.
  on Knowledge Representation and Reasoning, (KR '96)}, pp. 148--159, (1996).

\bibitem{fahle1}
T.~Fahle, S.~Schamberger, and M.~Sellmann, `Symmetry breaking', in {\em
  7th Int. Conf. on Principles and Practice of
  Constraint Programming (CP2001)}, pp. 93--107. 
  (2001).

\bibitem{ffhkmpwcp2002}
P.~Flener, A.~Frisch, B.~Hnich, Z.~Kiziltan, I.~Miguel, J.~Pearson, and
  T.~Walsh, `Breaking row and column symmetry in matrix models', in {\em 8th
  Int. Conf. on Principles and Practices of Constraint
  Programming (CP-2002)}. (2002).

\bibitem{fhkmwcp2002}
A.~Frisch, B.~Hnich, Z.~Kiziltan, I.~Miguel, and T.~Walsh, `Global constraints
  for lexicographic orderings', in {\em 8th Int. Conf. on
  Principles and Practices of Constraint Programming (CP-2002)}. 
  (2002).

\bibitem{fhkmwaij06}
A.~Frisch, B.~Hnich, Z.~Kiziltan, I.~Miguel, and T.~Walsh, `Propagation
  algorithms for lexicographic ordering constraints', {\em Artificial
  Intelligence}, {\bf 170}(10),  803--908, (2006).

\bibitem{sbds}
I.P. Gent and B.M. Smith, `Symmetry breaking in constraint programming', in
  {\em ECAI-2000}, pp. 599--603. 
  (2000).

\bibitem{csplib}
I.P. Gent and T.~Walsh, `{CSPLib}: a benchmark library for constraints',
  Technical report, Technical report APES-09-1999, (1999).
  \newblock A shorter version appears in Proc. of CP-99.

\bibitem{gskaaai98}
C.~Gomes, B.~Selman, and H.~Kautz, `Boosting combinatorial search through
  randomization', in {\em 15th National Conf. on Artificial
  Intelligence}, pp. 431--437. (1998).

\bibitem{hpsycp08}
D.~Heller, A.~Panda, M.~Sellmann, and J.~Yip, `Model restarts for structural
  symmetry breaking', in {\em 14th Int. Conf. on the Principles
  and Practice of Constraint Programming}, pp. 539--544, (2008).

\bibitem{hmmncp09}
S.~Huczynska, P.~McKay, I.~Miguel, and P.~Nightingale, `Modelling equidistant
  frequency permutation arrays: An application of constraints to mathematics',
  in {\em 15th Int. Conf. on the Principles and Practice of Constraint Programming}, pp. 50--64.  (2009).

\bibitem{gaplex}
C.~Jefferson, T.~Kelsey, S.~Linton, and K.~Petrie, `Gaplex: Generalised static
  symmetry breaking', in {\em 6th Int. Workshop on
  Symmetry in Constraint Satisfaction Problems, held alongside
  CP-06}, (2006).

\bibitem{llwycp07}
Y.-C. Law, J.~Lee, T.~Walsh, and J.~Yip, `Breaking symmetry of interchangeable
  variables and values', in {\em 13th Int. Conf. on Principles
  and Practices of Constraint Programming (CP-2007)}.  (2007).

\bibitem{llconstraints06}
Y.C. Law and J.M.H. Lee, `{Symmetry Breaking Constraints for Value Symmetries
  in Constraint Satisfaction}', {\em Constraints}, {\bf 11}(2--3),  221--267,
  (2006).

\bibitem{puget:Sym}
J.-F. Puget, `On the satisfiability of symmetrical constrained satisfaction
  problems', in {\em Proc. of ISMIS'93}, pp. 350--361.  (1993).

\bibitem{puget2003}
J-F. Puget, `Symmetry breaking using stabilizers', in {\em 9th
  Int. Conf. on Principles and Practice of Constraint Programming
  (CP2003)},  (2003).

\bibitem{getree}
C.~Roney-Dougal, I.~Gent, T.~Kelsey, and S.~Linton, `Tractable symmetry
  breaking using restricted search trees', in {\em ECAI-2004}.
  (2004).

\bibitem{ssat2001}
I.~Shlyakhter, `Generating effective symmetry-breaking predicates for search
  problems', in {\em Proc. of LICS workshop on Theory and Applications of
  Satisfiability Testing (SAT 2001)}, (2001).

\bibitem{wecai2006}
T.~Walsh, `Symmetry breaking using value precedence', in {\em ECAI-2006}, pp. 168--172. (2006).

\bibitem{wcp07}
T.~Walsh, `Breaking value symmetry', in {\em 13th Int. Conf. on
  Principles and Practices of Constraint Programming (CP-2007)}.
   (2007).

\bibitem{waaai2008}
T.~Walsh, `Breaking value symmetry', in {\em Proc. of the 23rd National
  Conf. on AI}, pp. 1585--1588. AAAI, (2008).

\end{thebibliography}


\begin{thebibliography}{10}

\bibitem{kn:Adams85}
L.~Adams, `{m-Step} preconditioned {Gradient} methods', {\em SIAM Journal of
  Scientific and Statistical Computing}, {\bf 6},  452--463, (1985).

\bibitem{kn:Atkin}
P.~Atkin.
\newblock Performance maximisation.
\newblock INMOS Technical Note 17.

\bibitem{kn:daCunha92b}
R.D. {da Cunha} and T.R. Hopkins, {\em The Parallel Solution of Partial
  Differential Equations on Transputer Networks},  96--109, Transputing for
  Numerical and Neural Network Applications, IOS Press, Amsterdam, 1992.
\newblock Also as Report No. 17/92, Computing Laboratory, University of Kent at
  Canterbury, U.K.

\bibitem{kn:daCunha92a}
R.D. {da Cunha} and T.R. Hopkins, {\em The Parallel Solution of Systems of
  Linear Equations using Iterative Methods on Transputer Networks},  1--13,
  Transputing for Numerical and Neural Network Applications, IOS Press,
  Amsterdam, 1992.
\newblock Also as Report No. 16/92, Computing Laboratory, University of Kent at
  Canterbury, U.K.

\bibitem{kn:deCarlini91}
U.~{de Carlini} and U.~Villano, {\em Transputers and parallel architectures --
  message-passing distributed systems}, Ellis Horwood, Chichester, 1991.

\bibitem{kn:Eisenstat81}
S.C. Eisenstat, `Efficient implementation of a class of preconditioned
  {Conjugate Gradient} methods', {\em SIAM Journal of Scientific and
  Statistical Computing}, {\bf 2},  1--4, (1981).

\bibitem{kn:Golub89}
G.H. Golub and C.F. {Van Loan}, {\em Matrix Computations}, Johns Hopkins
  University Press, Baltimore, 2nd edn., 1989.

\bibitem{kn:Johnson83}
O.G. Johnson, C.A. Micchelli, and G.~Paul, `Polynomial preconditioners for
  {Conjugate Gradient} calculations', {\em SIAM Journal of Numerical Analysis},
  {\bf 20},  362--376, (1983).

\bibitem{kn:Modi88}
J.J. Modi, {\em Parallel Algorithms and Matrix Computation}, Oxford University
  Press, Oxford, 1988.

\bibitem{kn:Saad85}
Y.~Saad, `Practical use of polynomial preconditionings for the {Conjugate
  Gradient} method', {\em SIAM Journal of Scientific and Statistical
  Computing}, {\bf 6},  865--881, (1985).

\bibitem{kn:Schofield89}
C.F. Schofield, {\em Optimising {FORTRAN} programs}, Ellis Horwood Publishing,
  Chichester, 1989.

\bibitem{kn:Smith85}
G.D. Smith, {\em Numerical Solution of Partial Differential Equations: Finite
  Difference Methods}, Oxford University Press, Oxford, 3rd edn., 1985.

\end{thebibliography}

\end{document}